\documentclass[11pt,a4paper]{article}
\usepackage[hyperref]{acl2021}
\usepackage{times}
\usepackage{latexsym}

\usepackage{microtype}

\usepackage{graphicx}
\usepackage{multicol}
\graphicspath{ {./images/} }

\aclfinalcopy % Uncomment this line for the final submission
\title{Zhestyatsky at SemEval-2021 Task 2: ReLU over Cosine Similarity for BERT Fine-tuning}

\author{Boris Zhestiankin \\
  Moscow Institute of Physics and Technology \\
  Moscow, Russia \\ \\
  \texttt{zhestyankin.ba@phystech.edu} \\\And
  Maria Ponomareva \\
  ABBYY \\
  HSE University \\
  Moscow, Russia \\
  \texttt{maria.ponomareva@abbyy.com} \\}

\date{}

\begin{document}
\maketitle
\begin{abstract}
This paper presents our contribution to SemEval-2021 Task 2: Multilingual and Cross-lingual Word-in-Context Disambiguation (MCL-WiC). Our experiments cover English (EN-EN) sub-track from the multilingual setting of the task. We experiment with several pre-trained language models and investigate an impact of different top-layers on fine-tuning. We find the combination of Cosine Similarity and ReLU activation leading to the most effective fine-tuning procedure. Our best  model results in accuracy  92.7\%, which is the fourth-best score in EN-EN sub-track.
\end{abstract}

\section{Introduction}
The increasing progress in Natural Language Processing is closely related with development of word representations. The context-independent word embeddings, such as Word2Vec ~\citep{Mikolov13} and fastText ~\citep{bojanowski2017enriching} brought the idea of measuring the relatedness of the meanings as the distance between the vectors encoding them. The introduction of the methods of pre-training context dependent embeddings, such as ELMo ~\citep{peters-etal-2018-deep}, ULMFit ~\citep{Howard}, and BERT ~\citep{devlin2018pretraining} made the next crucial breakthrough overcoming the shortcomings of previous methods to encode the meaning. Despite the fact that the primal objective of word embeddings is to encode the meaning of words, it is not obvious how to evaluate them directly. While common manner to examine the superiority of particular type of embeddings is to look at their performance on some downstream tasks, the more direct way to evaluate their ability to represent semantic is challenging.

SemEval-2021 Task 2: Multilingual and Cross-lingual Word-in-Context Disambiguation (MCL-WiC) ~\citep{martelli-etal-2021-mclwic} presents a new framework to evaluate embeddings. In this paper we present our contribution for the task. We explore the potential of different pre-trained context-dependent embeddings based on pre-trained language models. We find that the Cosine Similarity can produce fruitful results when used for fine-tuning the weights of the pre-trained models,  while adding linear layers to learn the similarity from the limited data leads to instant overfitting.

\section{Background}
The traditional approach to evaluate the ability of embeddings to catch the meaning of words is Word Sense Disambiguation (WSD) task ~\citep{Navigli09wordsense}.  WSD is defined as classification problem, when a given word is classificated between its predefined senses. WSD by design comes with an important limitation, being connected directly with predefined sense inventories such as WordNet\footnote{\url{https://wordnet.princeton.edu}} ~\citep{fellbaum2005}. 

The Word in Context (WiC) benchmark  ~\citep{pilehvar19} addresses these limitations. The task proposes a binary classification setting for English, when, given two sentences \textit{s\textsubscript{i}} and \textit{s\textsubscript{k}} and two words \textit{w\textsubscript{i}} and \textit{w\textsubscript{k}} in them,  the system needs to decide whether the word  \textit{w\textsubscript{i}}  in  \textit{s\textsubscript{i}} and \textit{w\textsubscript{k}} in \textit{s\textsubscript{k}} have same or different meanings.
The main advantage of WiC task is a possibility to expand its consideration  to the languages that lack such sense inventories.

MCL-WiC extends the WiC approach to new senses and new languages, covering data in five languages: Arabic, Chinese, English, French and Russian. The task provides data of two types: in the multilingual setting one needs to predict the label to the pair of sentences in one language (AR-AR, ZH-ZH, EN-EN, FR-FR, RU-RU sub-tracks), in the cross-lingual setting the first sentence is in English and the second one is in one of the four other considered languages (EN-AR, EN-ZH, EN-FR, EN-RU sub-tracks).

After preliminary experiments we decided to focus our efforts on the only sub-track with training data, namely the English sub-track from the multilingual setting.
Our solution\footnote{Source code, experiments, requirements and results can be found at \url{https://github.com/zhestyatsky/MCL-WiC}} is fourth placed in the EN-EN leaderboard with 92.7\% accuracy and is 0.6\% behind the winner.

\begin{figure*}[ht]
  \centering
  \includegraphics[width=0.63\textwidth]{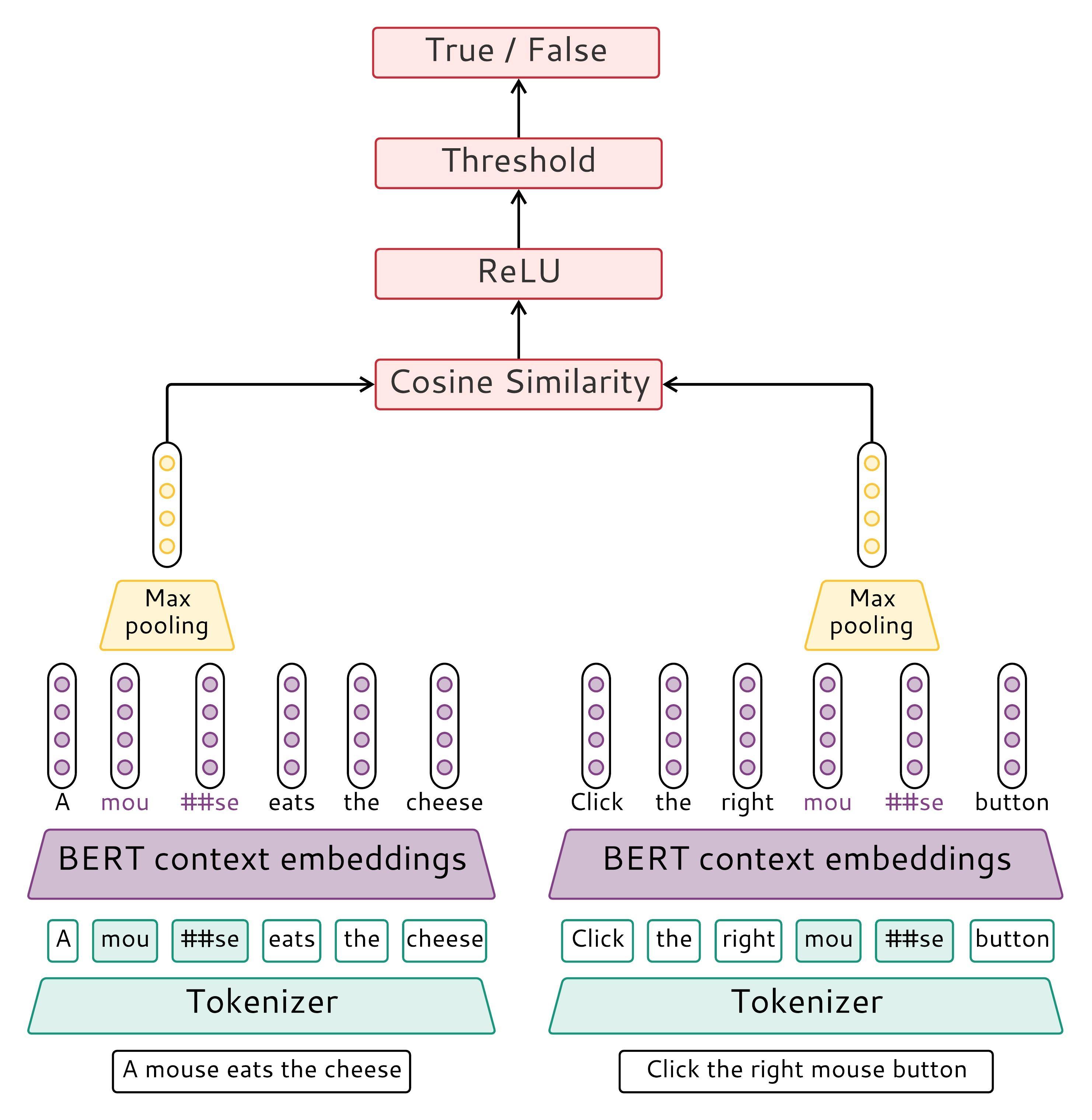}
  \caption{The scheme presents Cosine Similarity Architecture, which was used in the model achieving the best performance in our experiments.}
  \label{fig:diagram}
\end{figure*}

\section{System overview}
Approaching the task we conduct multiple experiments with a variety of architectures, however all of them are deeply based on contextual embeddings fine-tuning. For our experiments we use pre-trained embeddings from BERT  and XLM-RoBERTa ~\citep{conneau20} models and fine-tune them for our task.

\subsection{Target word embeddings}
Design of BERT and XLM-RoBERTa models assumes that text is first split to tokens and embeddings for these tokens are evaluated. Therefore we define our technique to obtain the embeddings, representing target words in the sentences.

For a single sentence we take embeddings of all sub-tokens corresponding to the target word in it and max pool them into one embedding. Repeating this procedure for both sentences in each pair we obtain two embeddings as the result: first — corresponding to target word in the first sentence and second — corresponding to target word in the second sentence.

\subsection{Multilayer Perceptron Architecture}

In our initial setup we build a system based on Multilayer Perceptron neural network. The purpose of this approach is to train the system to predict that target words have the same meaning in both sentences.

This model calculates embeddings of the target word in both sentences of the pair and concatenates them together, taking the result as an input layer. The model contains one hidden layer with 100 neurons, ReLU activation before it and an output layer, activated by sigmoid.

Interpreting the model output as the probability that target words have the same meaning in both of the sentences, we predict True if the output turns out to be greater than 0.5 or we predict False otherwise.

To enrich the knowledge of the model about the task we also experiment with a slightly different input, making use of [CLS] tokens. Each [CLS] token represents the whole sentence. Taking [CLS] tokens embeddings for each sentence in a pair we concatenate them together and afterwards concatenate the result with an input layer (consisting of target word embeddings concatenation) defined above. We use the resulting embedding as an input layer for our model and do not change other parameters in the setup.

\subsection{Cosine Similarity Architecture}
As an alternative to Multilayer Perceptron approach we define a Cosine Similarity approach, illustrated on Figure \ref{fig:diagram}. which proves to be our best system for the task. The purpose of this approach is to train the system to predict the probability that the target word has the same meaning in both sentences.

During training our system takes embeddings of the target word in each sentence in a pair and calculates Cosine Similarity between them. It activates the similarity through ReLU layer. The result value is considered the output of the model.

After the training is finished we have to make predictions, which is achieved by defining the probability threshold as a hyperparameter. In this way we predict True if the output of the model is greater than the threshold or False otherwise.

To maximize the accuracy of the model we calculate the probability threshold by building the Receiver Operating Characteristic (ROC) curve and choosing the value corresponding to the maximum difference between true positive and false positive rates.

We note that in this approach no new weights are introduced in contradistinction to Multilayer Perceptron approach. Therefore only pre-trained weights of BERT and XLM-RoBERTa models are fine-tuned.

To provide a comparison option for Cosine Similarity approach we also try applying sigmoid as an activation instead of ReLU.

\subsection{Datasets}
Speaking about the datasets for training and validation we fully utilize train and development English data provided by the competition organisers for the EN-EN sub-track. However, to achieve the best possible results we extend our train and development datasets with WiC dataset ~\citep{pilehvar19}\footnote{\url{https://pilehvar.github.io/wic/}}, included to SuperGLUE ~\citep{superglue} benchmark, for English sentence pairs.

We also conduct an experiment with our best model, using only default datasets provided by competition organisers. This experiment will be described at the end of Results section.

\section{Experimental setup}

In our setup we mix train and development data and split it randomly by unique lemmas in proportion 97.5\% to 2.5\%. Having 14680 samples in the first chunk and and 386 samples in the second chunk we use the first chunk for training and the second for validation.

During training, the data is processed by batches of size 8. Each sentence is split into 118 tokens maximum. In this way it is guaranteed that the longest sentence in the dataset is not going to be truncated.

Experiments with four different types of embeddings are conducted\footnote{\url{https://huggingface.co/transformers}}:

\begin{itemize}
    \item 
    \textit{bert-base-cased}: 12-layer, 768-hidden, 12-heads, 109M parameters;
    \item
    \textit{bert-large-cased}: 24-layer, 1024-hidden, 16-heads, 335M parameters;
    \item
    \textit{xlm-roberta-base}: $\sim$270M parameters with 12-layers, 768-hidden-state, 3072 feed-forward hidden-state, 8-heads;
    \item
    \textit{xlm-roberta-large}: $\sim$550M parameters with 24-layers, 1024-hidden-state, 4096 feed-forward hidden-state, 16-heads.
\end{itemize}

We train our models for a maximum of 8 epochs and define an early stopping criteria. Every half of epoch (after training on the half of all the batches) we check if the loss on validation dataset is decreasing. If the loss does not decrease for 2 checks in a row, we stop training.

In all our experiments we use Binary Cross Entropy Loss as the loss function and AdamW optimizer with a learning rate set to 1e-5.

To conduct experiments we use version 1.7.1 of PyTorch ~\citep{Torch} together with version 0.8.2 of torchvision\footnote{\url{https://github.com/pytorch/vision}} and version 0.8.1 of torchtext\footnote{\url{https://github.com/pytorch/text}}, version 1.1.6 of PyTorch Lightning\footnote{\url{https://github.com/PyTorchLightning/pytorch-lightning}} framework and version 4.2.2 of HuggingFace's Transformers ~\citep{wolf20}. From the latter we obtain BERT and XLM-RoBERTa model implementations.

As we define a probability threshold as a hyperparameter in Cosine Similarity approach, we provide its values for all experimental configurations in the Table \ref{tab:my-table}.

\begin{table}[ht]
\centering
\begin{tabular}{|c|c|c|}
\hline
\textbf{embeddings}        & \textbf{activation}               & \textbf{threshold} \\ \hline
\textit{xlm-roberta-large} & \textbf{sigmoid} & 0.680         \\ \cline{1-1} \cline{3-3} 
\textit{xlm-roberta-base}  &                                   & 0.632         \\ \cline{1-1} \cline{3-3} 
\textit{bert-large-cased}  &                                   & 0.609         \\ \cline{1-1} \cline{3-3} 
\textit{bert-base-cased}   &                                   & 0.678         \\ \hline
\textit{xlm-roberta-large} & \textbf{ReLU}    & 0.638         \\ \cline{1-1} \cline{3-3} 
\textit{xlm-roberta-base}  &                                   & 0.642         \\ \cline{1-1} \cline{3-3} 
\textit{bert-large-cased}  &                                   & 0.519         \\ \cline{1-1} \cline{3-3} 
\textit{bert-base-cased}   &                                   & 0.509         \\ \hline
\end{tabular}
\caption{Probability thresholds for Cosine Similarity Architecture. Abbreviations used: \textbf{activation} stands for \textit{activation function} used, \textbf{threshold} stands for \textit{probability threshold} of the model.}
\label{tab:my-table}
\end{table}

\section{Results}

In the Table \ref{fine linear} the results of the fine-tuning of language models with Multilayer Perceptron on top are presented. During the experiments we found out that for this dataset not only additional linear layers can not learn to measure the distance effectively, but they lead to overfitting in a few epochs. It is seen by the number of the passed epochs before the early stopping.

As [CLS] token is designed to accumulate sentence meaning we expected it to make the representations for each instance in a pair more complete. The results in the Table \ref{fine linear} show that the usage of [CLS] tokens give a moderate improvement to all models except for one with \textit{xlm-roberta-large} embeddings.

\begin{table}[ht]
\begin{tabular}{|c|c|c|c|c|}
\hline
\textbf{embed} & \textbf{add cls} & \textbf{epochs} & \textbf{val}  & \textbf{test}  \\
\hline
\textbf{XLMR-l} & \textbf{yes} & 2.5                                 & 0.585          & 0.579          \\
\textbf{XLMR-b}  &                                  & 2.5                                 & 0.580           & 0.580           \\
\textbf{BERT-l}    &                                  & 3.5                                 & 0.585          & 0.548          \\
\textbf{BERT-b}     &                                  & 2.5                                 & 0.588          & 0.565          \\
\hline
\textbf{XLMR-l} & \textbf{no}    & 2                                   & 0.484          & 0.519          \\
\textbf{XLMR-b}  &                                  & 2.5                                 & 0.590           & 0.611          \\
\textbf{BERT-l}    &                                  & 3                                   & 0.598          & 0.583          \\
\textbf{BERT-b}     &                                  & 2.5                                 & \textbf{0.601} & \textbf{0.592} \\
\hline
\end{tabular}
\caption{\label{fine linear}
Accuracy of models with Multilayer Perceptron Architecture. Abbreviations used: \textbf{embed} stands for \textit{embeddings},  \textbf{add cls} defines if \textit{[CLS] token embedding was used}, \textbf{val} stands for \textit{accuracy on validation dataset}, \textbf{test} stands for \textit{accuracy on test dataset}. We refer to \textit{xlm-roberta-large} as \textbf{XLMR-l}, to  \textit{xlm-roberta-base} as \textbf{XLMR-b}, to  \textit{bert-large-cased} as \textbf{BERT-l} and to \textit{bert-base-cased} as \textbf{BERT-b}.
}
\end{table}

Pre-trained language models, like BERT  and XLM-RoBERTa, have the property of associating close vectors with similar words. Therefore to provide a baseline for the model described in Cosine Similarity approach we measure the accuracy of it without additional fine-tuning. Due to the technique used to evaluate the probability thresholds, the accuracies for configurations with different activations are identical in this case. Accuracies for different embeddings and thresholds for sigmoid and ReLU activations can be found in Table \ref{no fine}. Viewing the results on validation dataset we can estimate the quality of the approach and the results on test dataset confirm its relevance. Best accuracy on validation dataset is provided by \textit{bert-large-cased} embeddings. In addition, the thresholds in Table \ref{no fine} show how differently the vector spaces are arranged for BERT  and XLM-RoBERTa models: for the second, a threshold of about 0.99 distinguishes vectors of words with different meanings from words with the same meanings.

\begin{table}[ht!]
\begin{tabular}{|c|c|c|c|c|}
\hline
     \textbf{embed}           & \textbf{\begin{tabular}[c]{@{}c@{}}sigm\\  thld \end{tabular}} & \textbf{\begin{tabular}[c]{@{}c@{}}ReLU \\ thld \end{tabular}} & \textbf{val}  & \textbf{test} \\
\hline
\textbf{XLMR-l} & 0.73                                                                  & 0.995                                                              & 0.645         & 0.659         \\
\textbf{XLMR-b} & 0.72                                                                  & 0.994                                                              & 0.666         & 0.719         \\
\textbf{BERT-l} & 0.66                                                                  & 0.64                                                               & \textbf{0.710} & \textbf{0.780} \\
\textbf{BERT-b} & 0.69                                                                  & 0.77                                                               & 0.690          & 0.780    \\
\hline
\end{tabular}
\caption{\label{no fine}
Accuracy of models with Cosine Similarity Architecture without fine-tuning. Abbreviations used: \textbf{embed} stands for \textit{embeddings},  \textbf{sigm thld} stands for \textit{probability threshold of model using sigmoid activation}, \textbf{ReLU thld} stands for \textit{probability threshold of model using ReLU activation}, \textbf{val} stands for \textit{accuracy on validation dataset}, \textbf{test} stands for \textit{accuracy on test dataset}. As models are not fine-tuned, accuracies on validation and test datasets are independent of the activation function. We refer to \textit{xlm-roberta-large} as \textbf{XLMR-l}, to  \textit{xlm-roberta-base} as \textbf{XLMR-b}, to  \textit{bert-large-cased} as \textbf{BERT-l} and to \textit{bert-base-cased} as \textbf{BERT-b}.
}
\end{table}

\ 

Finally, Table \ref{fine cosine}  presents results of the experimental setup when  the language models  are fine-tuned using Cosine Similarity measure. It is worth mentioning that in such a setup there are no additional weights and only  the layers of the language model are changing. It can be seen that such an architecture allows th model not to overfit for longer epochs. 

\begin{table}[ht!]
\begin{tabular}{|c|c|c|c|c|}
\hline
\textbf{embed} &     \textbf{activ} &\textbf{epochs} &\textbf{val}   &\textbf{test}  \\
\hline
\textbf{XLMR-l}     & {\textbf{sigm}} & 6                                   & 0.661          & 0.748          \\
\textbf{XLMR-b}     &                                   & 3                                   & 0.679          & 0.746          \\
\textbf{BERT-l}     &                                   & 3                                   & 0.728          & 0.823          \\
\textbf{BERT-b}     &                                   & 3                                   & 0.676          & 0.727          \\
\hline
\textbf{XLMR-l}     & {\textbf{ReLU}}    & 5.5                                 & 0.785          & 0.876          \\
\textbf{XLMR-b}     &                                   & 2                                   & 0.730           & 0.769          \\
\textbf{BERT-l}     &                                   & \textbf{4.5}                        & \textbf{0.808} & \textbf{0.927} \\
\textbf{BERT-b}     &                                   & 4                                   & 0.790           & 0.889        \\
\hline
\end{tabular}
\caption{\label{fine cosine}
Accuracy of models with Cosine Similarity Architecture. Abbreviations used: \textbf{embed} stands for \textit{embeddings}, \textbf{activ} stands for the \textit{activation function} used, \textbf{sigm} stands for \textit{sigmoid activation function}, \textbf{val} stands for \textit{accuracy on validation dataset}, \textbf{test} stands for \textit{accuracy on test dataset}. We refer to \textit{xlm-roberta-large} as \textbf{XLMR-l}, to  \textit{xlm-roberta-base} as \textbf{XLMR-b}, to  \textit{bert-large-cased} as \textbf{BERT-l} and to \textit{bert-base-cased} as \textbf{BERT-b}.
}
\end{table}

While conducting the experiments, we judged the models by their performance on the validation dataset, not being able to check how representative it is. According to the obtained scores, the validation dataset is representative enough and is more challenging for the models than the test dataset.

To provide a convenient report we conduct an experiment with our best model (using \textit{bert-large-cased} embeddings together with Cosine Similarity Architecture, using ReLU activation), which only uses data provided by organisers. We perform no further processing with the data and use it as is: train dataset is used for training and development dataset for validation. Being trained for 4 epochs the model in the experiment demonstrates 0.886 accuracy on validation dataset and \textbf{0.913} accuracy on test dataset. This result shows that using additional data leads to better performance.

\section{Error analysis}
Our best model leads to accuracy 92.7\%. It means that our model has erroneously labeled 73 sentences in the 1000-sentence testset.
The error analysis revealed that our model is not biased towards one or another class, it produced 37 false negative predictions and 36 false positives predictions. The next observation is related to the construction feature of the dataset.
The dataset is organized in the following manner: for each combination of lemma and POS-tag there are two instances in the dataset. All three possible combinations of labels are presented, with prevalent case when one pair is labeled False and second True. The peculiarity of the dataset is that both instances have the same first sentence. We found that 20 out of 73 errors have these repeating first sentence.  
In other words, if the model produces incorrect prediction for one instance for lemma it tends to make a mistake for the second instance in the dataset. Due to the described peculiarity of the data, we can not speculate that certain lemma is a stumbling block for the model or it is just a context of the first sentence, that for example differs by genre or thematically from second sentence and complicates the prediction.
The manual analysis of the errors has not revealed instances that could be considered hard and unclear for human assessment.

In order to reveal objectively hard instances among the errors of the best model, we have intersected the mislabeled pairs for all the models fine-tuned with Cosine Similarity. The intersection indicated that all but two instances were predicted correctly by at least one of the models. We can conclude that no objectively hard instances were presented in the erroneously labeled pairs by the best model. Additionally, the possible conclusion could be that an ensemble of our models could result in even more powerful solution for the task. 

\section{Conclusion}

We have provided an overview of different approaches to fine-tune pre-trained language models for the task that is naturally suitable for them -- detecting the distance between representations of the words.

We have showed that, for the data of given amount and type, learning distance between words in context with Multilayer Perceptron neural network is not applicable and generally leads to overfitting.

Using Cosine Similarity to predict probability during pre-trained embeddings fine-tuning leads to much more promising results, when activated with ReLU layer.

\bibliographystyle{acl_natbib}
\bibliography{semeval}

\begin{thebibliography}{13}
\expandafter\ifx\csname natexlab\endcsname\relax\def\natexlab#1{#1}\fi

\bibitem[{Bojanowski et~al.(2017)Bojanowski, Grave, Joulin, and
  Mikolov}]{bojanowski2017enriching}
Piotr Bojanowski, Edouard Grave, Armand Joulin, and Tomas Mikolov. 2017.
\newblock Enriching word vectors with subword information.
\newblock \emph{Transactions of the Association for Computational Linguistics},
  5:135--146.

\bibitem[{Conneau et~al.(2020)Conneau, Khandelwal, Goyal, Chaudhary, Wenzek,
  Guzm{\'a}n, Grave, Ott, Zettlemoyer, and Stoyanov}]{conneau20}
Alexis Conneau, Kartikay Khandelwal, Naman Goyal, Vishrav Chaudhary, Guillaume
  Wenzek, Francisco Guzm{\'a}n, Edouard Grave, Myle Ott, Luke Zettlemoyer, and
  Veselin Stoyanov. 2020.
\newblock \href {https://doi.org/10.18653/v1/2020.acl-main.747} {Unsupervised
  cross-lingual representation learning at scale}.
\newblock In \emph{Proceedings of the 58th Annual Meeting of the Association
  for Computational Linguistics}, pages 8440--8451, Online. Association for
  Computational Linguistics.

\bibitem[{Devlin et~al.(2018)Devlin, Chang, Lee, and
  Toutanova}]{devlin2018pretraining}
Jacob Devlin, Ming-Wei Chang, Kenton Lee, and Kristina Toutanova. 2018.
\newblock \href {http://arxiv.org/abs/1810.04805} {Bert: Pre-training of deep
  bidirectional transformers for language understanding}.
\newblock Cite arxiv:1810.04805Comment: 13 pages.

\bibitem[{Fellbaum(2005)}]{fellbaum2005}
Christiane Fellbaum. 2005.
\newblock \href {http://wordnet.princeton.edu/} {Wordnet and wordnets}.
\newblock In \emph{Encyclopedia of Language and Linguistics}, pages 665--670,
  Oxford. Elsevier.

\bibitem[{Howard and Ruder(2018)}]{Howard}
Jeremy Howard and Sebastian Ruder. 2018.
\newblock \href {http://arxiv.org/abs/1801.06146} {Fine-tuned language models
  for text classification}.
\newblock \emph{CoRR}, abs/1801.06146.

\bibitem[{Martelli et~al.(2021)Martelli, Kalach, Tola, and
  Navigli}]{martelli-etal-2021-mclwic}
Federico Martelli, Najla Kalach, Gabriele Tola, and Roberto Navigli. 2021.
\newblock {S}em{E}val-2021 {T}ask 2: {M}ultilingual and {C}ross-lingual
  {W}ord-in-{C}ontext {D}isambiguation ({MCL}-{W}i{C}).
\newblock In \emph{Proceedings of the Fifteenth Workshop on Semantic Evaluation
  (SemEval-2021)}.

\bibitem[{Mikolov et~al.(2013)Mikolov, Sutskever, Chen, Corrado, and
  Dean}]{Mikolov13}
Tomas Mikolov, Ilya Sutskever, Kai Chen, Greg Corrado, and Jeffrey Dean. 2013.
\newblock \href
  {https://papers.nips.cc/paper/5021-distributed-representations-of-words-and-phrases-and-their-compositionality.pdf}
  {Distributed representations of words and phrases and their
  compositionality}.
\newblock In \emph{Neural and Information Processing System (NIPS)}.

\bibitem[{Navigli(2009)}]{Navigli09wordsense}
Roberto Navigli. 2009.
\newblock Word sense disambiguation: a survey.
\newblock \emph{ACM COMPUTING SURVEYS}, 41(2):1--69.

\bibitem[{Paszke et~al.(2019)Paszke, Gross, Massa, Lerer, Bradbury, Chanan,
  Killeen, Lin, Gimelshein, Antiga, Desmaison, Kopf, Yang, DeVito, Raison,
  Tejani, Chilamkurthy, Steiner, Fang, Bai, and Chintala}]{Torch}
Adam Paszke, Sam Gross, Francisco Massa, Adam Lerer, James Bradbury, Gregory
  Chanan, Trevor Killeen, Zeming Lin, Natalia Gimelshein, Luca Antiga, Alban
  Desmaison, Andreas Kopf, Edward Yang, Zachary DeVito, Martin Raison, Alykhan
  Tejani, Sasank Chilamkurthy, Benoit Steiner, Lu~Fang, Junjie Bai, and Soumith
  Chintala. 2019.
\newblock \href
  {http://papers.neurips.cc/paper/9015-pytorch-an-imperative-style-high-performance-deep-learning-library.pdf}
  {Pytorch: An imperative style, high-performance deep learning library}.
\newblock In H.~Wallach, H.~Larochelle, A.~Beygelzimer, F.~d\textquotesingle
  Alch\'{e}-Buc, E.~Fox, and R.~Garnett, editors, \emph{Advances in Neural
  Information Processing Systems 32}, pages 8024--8035. Curran Associates, Inc.

\bibitem[{Peters et~al.(2018)Peters, Neumann, Iyyer, Gardner, Clark, Lee, and
  Zettlemoyer}]{peters-etal-2018-deep}
Matthew Peters, Mark Neumann, Mohit Iyyer, Matt Gardner, Christopher Clark,
  Kenton Lee, and Luke Zettlemoyer. 2018.
\newblock \href {https://doi.org/10.18653/v1/N18-1202} {Deep contextualized
  word representations}.
\newblock In \emph{Proceedings of the 2018 Conference of the North {A}merican
  Chapter of the Association for Computational Linguistics: Human Language
  Technologies, Volume 1 (Long Papers)}, pages 2227--2237, New Orleans,
  Louisiana. Association for Computational Linguistics.

\bibitem[{Pilehvar and Camacho-Collados(2019)}]{pilehvar19}
Mohammad~Taher Pilehvar and Jose Camacho-Collados. 2019.
\newblock \href {https://doi.org/10.18653/v1/N19-1128} {{W}i{C}: the
  word-in-context dataset for evaluating context-sensitive meaning
  representations}.
\newblock In \emph{Proceedings of the 2019 Conference of the North {A}merican
  Chapter of the Association for Computational Linguistics: Human Language
  Technologies, Volume 1 (Long and Short Papers)}, pages 1267--1273,
  Minneapolis, Minnesota. Association for Computational Linguistics.

\bibitem[{Wang et~al.(2019)Wang, Pruksachatkun, Nangia, Singh, Michael, Hill,
  Levy, and Bowman}]{superglue}
Alex Wang, Yada Pruksachatkun, Nikita Nangia, Amanpreet Singh, Julian Michael,
  Felix Hill, Omer Levy, and Samuel~R. Bowman. 2019.
\newblock Super{GLUE}: A stickier benchmark for general-purpose language
  understanding systems.
\newblock \emph{arXiv preprint 1905.00537}.

\bibitem[{Wolf et~al.(2020)Wolf, Debut, Sanh, Chaumond, Delangue, Moi, Cistac,
  Rault, Louf, Funtowicz, Davison, Shleifer, von Platen, Ma, Jernite, Plu, Xu,
  Le~Scao, Gugger, Drame, Lhoest, and Rush}]{wolf20}
Thomas Wolf, Lysandre Debut, Victor Sanh, Julien Chaumond, Clement Delangue,
  Anthony Moi, Pierric Cistac, Tim Rault, Remi Louf, Morgan Funtowicz, Joe
  Davison, Sam Shleifer, Patrick von Platen, Clara Ma, Yacine Jernite, Julien
  Plu, Canwen Xu, Teven Le~Scao, Sylvain Gugger, Mariama Drame, Quentin Lhoest,
  and Alexander Rush. 2020.
\newblock \href {https://doi.org/10.18653/v1/2020.emnlp-demos.6} {Transformers:
  State-of-the-art natural language processing}.
\newblock In \emph{Proceedings of the 2020 Conference on Empirical Methods in
  Natural Language Processing: System Demonstrations}, pages 38--45, Online.
  Association for Computational Linguistics.

\end{thebibliography}

%\appendix

\end{document}